\documentclass{article}
\usepackage{subcaption}

\usepackage[preprint]{neurips_2024}
\usepackage{amsmath}
\usepackage{booktabs}

\usepackage[utf8]{inputenc} 
\usepackage[T1]{fontenc}    
\usepackage{hyperref}       
\usepackage{url}            
\usepackage{booktabs}       
\usepackage{wrapfig}        
\usepackage{graphicx}      
\usepackage{amsfonts}       
\usepackage{nicefrac}       
\usepackage{microtype}      
\usepackage{xcolor}         

\title{Heterogeneous Low-Bandwidth Pre-Training of LLMs}

\author{\textbf{Yazan Obeidi}\textsuperscript{1}\thanks{Correspondence to \texttt{yazan@tplr.ai,amir@tplr.ai}}
  \quad
  \textbf{Amir Sarfi}\textsuperscript{1} \quad
  \textbf{Joel Lidin}\textsuperscript{1} \quad
  \textbf{Paul Janson}\textsuperscript{2} \quad
  \textbf{Eugene Belilovsky}\textsuperscript{2} \\
  \textsuperscript{1}Covenant AI \quad
  \textsuperscript{2}Mila, Concordia University 
}

\newcommand{\Topk}{\textsc{Top-}\kappa}

\begin{document}

\maketitle

\begin{abstract}
Pre-training large language models (LLMs) increasingly requires distributed compute, yet bandwidth constraints make it difficult to scale beyond well-provisioned datacenters—especially when model parallelism forces frequent, large inter-device communications. We study whether SparseLoCo, a low-communication data parallel method based on infrequent synchronization and sparse pseudo-gradient exchange, can be combined with low-bandwidth pipeline model parallelism via activation and activation-gradient compression. We introduce a heterogeneous distributed training framework where some participants host full replicas on high-bandwidth interconnects, while resource-limited participants are grouped to jointly instantiate a replica using pipeline parallelism with subspace-projected inter-stage communication. To make the recently introduced subspace pipeline compression compatible with SparseLoCo, we study a number of adaptations. Across large-scale language modeling experiments (178M–1B parameters) on standard pretraining corpora, we find that activation compression composes with SparseLoCo at modest cost, while selective (heterogeneous) compression consistently improves the loss–communication tradeoff relative to compressing all replicas—especially at aggressive compression ratios. 
These results suggest a practical path to incorporating  low-bandwidth model parallelism and heterogeneous participants into LLM pre-training.
\end{abstract}

\section{Introduction}

The size and scale of foundation models and large language models (LLMs) pre-training have increased drastically in recent years. This has resulted in the centralization of training into a handful of entities with large enough datacenters. In order to permit distributed training of LLMs that goes beyond a single datacenter, low-bandwidth distributed training for these models has become a topic of increasing interest \cite{diloco,SparseLoCo,pmlr-v202-wang23t}.

A large body of work has focused on techniques for reducing the cost of low-bandwidth data parallelism \cite{adaptive_fed_opt}. This includes techniques for reducing the frequency of communication \cite{adaptive_fed_opt,local_sgd,diloco,muloco}, the size of communicated messages \cite{pmlr-v202-wang23t,demo}, and the overlap of communication and computation \cite{douillard2025streaming,acco}. However, as the size of models grows, the need to train them across multiple accelerators through model parallelism techniques becomes critical. On the other hand, in many cases of interest such as training with a large number of global participants, ensuring a single training replica is contained within a well-interconnected set of accelerators becomes prohibitive. Indeed, in many practical cases, this constraint can significantly restrict the type of hardware that can be used and who can participate (e.g. only those with large resources). To address this question, several recent works consider low-bandwidth model parallelism \cite{SWARM,protocol_models, singh2025model}. Most relevant to our work, Ramasinghe et al. \cite{protocol_models} have shown that in the setting of standard data parallel methods such as training with the AdamW optimizer, an efficient activation compression method can be combined with pipeline parallelism to significantly reduce the bandwidth constraints of model parallelism. 

In this work, we consider the recently introduced data parallel method, SparseLoCo~\cite{SparseLoCo}, which is able to achieve data parallel performance with drastically reduced communication cost. We ask whether it can be combined with low-bandwidth model parallel techniques, and specifically the recent approach of Subspace Compression~\cite{protocol_models}. We demonstrate the necessary ingredients for Subspace compression required to closely maintain performance of SparseLoCo while achieving the ability to break down model replicas across accelerators with low-bandwidth between them. Subsequently, we propose a heterogeneous scheme in which participants with well interconnected accelerators host full replicas using standard parallelization without subspace projection, while resource-limited participants are grouped to jointly instantiate a replica via pipeline parallelism with subspace-compressed inter-stage communications over the Internet. This enables otherwise unusable nodes to contribute to the training, while applying compression only where bandwidth is the bottleneck.

Our overall contributions are (a) we demonstrate that SparseLoCo---a highly efficient local optimization and gradient compression method---can be combined with model-level activation compression and provide necessary adaptations for this, and (b) 
we propose a heterogeneous method that applies subspace compression selectively, and we characterize its behavior and scalability, demonstrating it can lead to both a practical use case and better performance.

\section{Method}
\label{sec:method}
We propose a communication-efficient framework for training over \emph{heterogeneous} hardware in distributed training of LLMs which we call Heterogeneous SparseLoCo. Each participant contributes a compute node with heterogeneous resources and communicates with other participants over the Internet, typically under limited-bandwidth. In this setting, (i) participants with well-connected multi-accelerator clusters can host full model replicas, and (ii) limited-compute participants are grouped to jointly instantiate a replica. We then use the recently introduced SparseLoCo~\cite{SparseLoCo} to enable infrequent synchronizations and sparse gradient compression across replicas, and we employ the gradient and activation compression method from Subspace Networks~\cite{protocol_models} to reduce within-replica communications of bandwidth-limited participant groups.

\begin{figure}
    \centering
    \includegraphics[width=\linewidth,trim={0cm 6cm 0cm 0cm},clip]{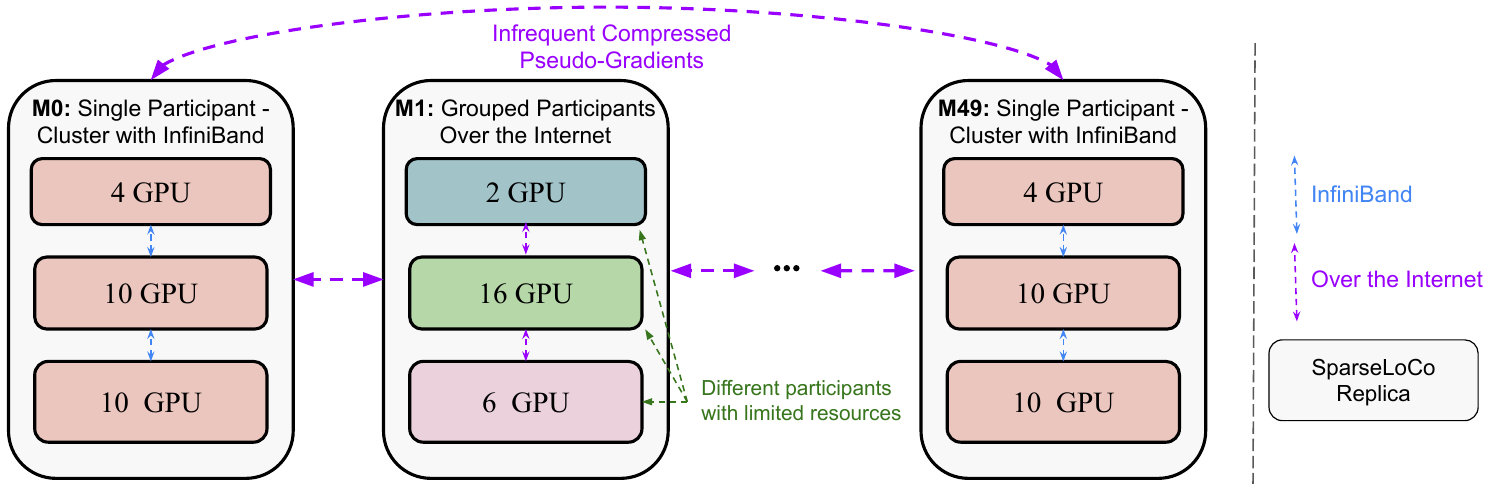}
    \caption{Illustration of heterogeneous training proposed in this work. Across the data parallel axis, we train with $M$ SparseLoCo replicas that perform $H$ local steps before synchronizing using compressed pseudo-gradient communication. A replica can be hosted by a single high-bandwidth cluster (M0 and M49), or by clustering a group of resource-limited participants that collectively form a replica via pipeline parallelism with compressed inter-stage communication (M1). Within each participant, any standard parallelism may be used. This framework allows effective training of large language models with low-bandwidth, heterogeneous resources.}
    \label{fig:mixedreplica}
\end{figure}

Specifically, we consider training a transformer architecture~\cite{transformer,llama1} that can be partitioned into $S$ transformer pipeline stages. The distributed system is organized into $M$ SparseLoCo replicas, where each replica ($m$) corresponds either to a single high-bandwidth cluster or to a group of smaller contributors. Within each replica, the model can be parallelized using standard schemes (e.g., tensor parallelism, pipeline parallelism, distributed data parallelism, etc.), and each pipeline stage may itself employ additional parallelism (e.g., tensor parallelism or sharding). This mixed-replica setting is illustrated in Figure~\ref{fig:mixedreplica}.

\subsection{Background: SparseLoCo}
\label{sec:dp_compress}

We reduce communication overhead in data parallelism using SparseLoCo~\cite{SparseLoCo}, which extends DiLoCo~\cite{diloco,adaptive_fed_opt} with pseudo-gradient compression and the removal of outer momentum for an error feedback mechanism.
Specifically, we consider  $H$ consecutive local optimization steps on each replica before synchronization, reducing communication frequency by a factor of $H$ while performing compression of the pseudo-gradients by selecting top$\kappa$ largest-magnitude values. For stage $s$ on replica $m$ at outer step $t$, we perform the following inner step $H$ times:
\begin{align}
\theta^{(t)}_{s,m} &\leftarrow \operatorname{InnerOpt}(\theta^{(t-1)}_{s}; \mathcal{D}_m), \quad \forall m \in \{1, \ldots, M\} \label{eq:inner_opt} 
\end{align}
Here, the $\operatorname{InnerOpt}$ is typically the classical AdamW optimizer~\cite{adam,adamw} with the local data being denoted as $\mathcal{D}_m$. The above inner step is repeated $H$ times and subsequently the following global model update is performed:

\begin{align}
&\Delta^{(t)}_{s,m} \leftarrow \theta^{(t-1)}_{s} - \theta^{(t)}_{s,m} \notag\\
&e^{(t)}_{s,m} \leftarrow \beta \, e^{(t-1)}_{s,m} + \Delta^{(t)}_{s,m} \notag \\
&\hat{\Delta}^{(t)}_{s,m} \leftarrow Q\left(\Topk\left(e^{(t)}_{s,m}\right)\right) \notag \\
&e^{(t+1)}_{s,m} \leftarrow e^{(t)}_{s,m} - \hat{\Delta}^{(t)}_{s,m} \notag \\
&\bar{\Delta}_s^{(t)} \leftarrow \frac{1}{M} \sum_{m=1}^{M} \hat{\Delta}^{(t)}_{s,m}, \quad 
\theta^{(t)}_{s} \leftarrow \theta^{(t-1)}_{s} - \eta \, \bar{\Delta}_s^{(t)} \label{eq:sparse_update}
\end{align}

The $\Topk$ operator retains only the $\kappa$ largest-magnitude elements per chunk, 
which are then quantized via $Q(\cdot)$, achieving sparsity while ensuring convergence 
through error feedback. The error accumulator $e^{(t)}_{s,m}$ preserves discarded 
pseudo-gradient information across outer steps.

\subsection{Background: Communication-Efficient Pipeline Parallelism}
\label{sec:pp_compress}

With pipeline parallelism, communication occurs at stage boundaries in both the forward and backward passes:
\begin{itemize}
    \item \textit{Forward pass:} Activation tensors $X_s \in \mathbb{R}^{b \times L \times d_{\text{model}}}$ flow from stage $s$ to stage $s+1$
    \item \textit{Backward pass:} Gradient tensors $\nabla_{X_s} \mathcal{L} \in \mathbb{R}^{b \times L \times d_{\text{model}}}$ flow from stage $s+1$ to stage $s$
\end{itemize}
where $d_{\text{model}}$ denotes the hidden dimension, $b$ the micro-batch size, and $L$ the sequence length.

To allow groups of resource-limited participants to form a replica, we adopt the subspace projection method from Subspace Networks~\cite{protocol_models} to compress inter-stage activations and activation-gradients, reducing communication overhead of bandwidth-limited nodes. In this regime, each participant takes on a pipeline stage with over the Internet inter-stage communication. 

As shown in Subspace Networks~\cite{protocol_models}, transformer activations $X^{(\ell)} \in \mathbb{R}^{b \times L \times d}$ at layer $\ell$ can be decomposed as:
\begin{equation*}
X^{(\ell+1)} = \underbrace{\sum_{i=1}^{\ell} \left( X^{(i)}_{\text{hidden}} W_{p_2}^{(i)} + X^{(i)}_{\text{attn}} W_{p_1}^{(i)} \right)}_{\text{low rank components}} + \underbrace{X^{(0)}}_{\text{high rank components}}
\end{equation*}
where $X^{(0)} = \text{TokenEmbed}(\mathbf{x}) + \text{PosEmbed}$. When the rows of the projection matrices $W_{p_1}$, $W_{p_2}$ are constrained to the column space of an orthonormal matrix $U \in \mathbb{R}^{d \times k}$ ($k \ll d$), the residual activations $\hat{X}^{(\ell)} = X^{(\ell)} - X^{(0)}$ lie entirely within the $k$-dimensional subspace $\mathcal{S} = \text{Col}(U)$.

Following the Subspace Networks framework ~\cite{protocol_models}, we leverage low-dimensional subspace projection to reduce communication overhead between pipeline stages. A randomly initialized matrix \( U \in \mathbb{R}^{d \times k} \) is orthonormalized via QR factorization to serve as the projection basis. Since residual activations occupy the low-dimensional subspace spanned by $\text{Col}(U)$, they can be projected without information loss. Additionally, \cite{protocol_models} suggests to further decompose embedding matrix into fixed high rank ($T_\perp[\mathbf{x}]$) and learnable low rank components, eliminating the need to transmit embedding matrices across stages. During the forward pass, only the subspace projections are transmitted:
\begin{equation*}
\tilde{X}_s = \left(X_s - T_\perp[\mathbf{x}] - \text{PosEmbed}\right) U \in \mathbb{R}^{b \times L \times k}
\end{equation*}
with reconstruction performed via:
\begin{equation*}
\hat{X}_s = \tilde{X}_s U^\top + T_\perp[\mathbf{x}] + \text{PosEmbed}
\end{equation*}
For backward gradients, the same projection is applied:
\begin{equation*}
(\nabla_{X_s} \mathcal{L})_{\text{compressed}} = \nabla_{X_s} \mathcal{L} \cdot U
\end{equation*}

\subsection{Heterogeneous Training Configuration}
\label{sec:heterogeneous}
As shown in Figure~\ref{fig:mixedreplica}, we investigate training scenarios involving heterogeneous clusters, where replicas can have accelerators with varying quality. Specifically, we consider environments containing both high-bandwidth interconnects (e.g. InfiniBand clusters) and lower-bandwidth networks (e.g., Internet). We propose to selectively apply pipeline compression based on interconnect bandwidth: well-connected accelerator groups perform standard parallelism without subspace projection, eliminating information loss from compression, while poorly-connected groups employ the Subspace Network~\cite{protocol_models} compression scheme (Section~\ref{sec:pp_compress}) to maintain training throughput despite bandwidth constraints. This heterogeneous configuration still operates within the SparseLoCo~\cite{SparseLoCo} framework (Section~\ref{sec:dp_compress}). The stage level compression decisions apply during the $H$ inner optimization steps, while outer synchronization aggregates updates globally across all worker groups, regardless of their internal communication strategy. 

We demonstrate empirically that this heterogeneous strategy, where some replicas do not utilize activation compression, can mitigate performance degradation that arises from activation compression in replicas that use it.  

Practically, this allows for large training runs where entire datacenters can act as SparseLoCo replicas, while other replicas aggregate more poorly connected, but lower memory budget accelerators. Notably, accelerators representing different stages can be grouped based on their local proximity or bandwidth. Furthermore, datacenters with limited internal bandwidth (e.g. with cohosted machines) can utilize subspace compression while participating alongside high-bandwidth datacenters. 

 \subsection{Token Embeddings and Heterogeneous Setting} \label{sec:token}
 Token embeddings require special care in our mixed setting, where some replicas train with uncompressed parallelization while others apply subspace compression. As previously mentioned in Section \ref{sec:pp_compress}, we decompose the embedding table as $\mathrm{TE}=T_S+T_\perp$, where $T_S$ is constrained to the compression subspace $\mathcal{S}$ and $T_\perp$ stores the high-rank component that is shared among all pipeline stages. In the uniform setting, where all replicas perform subspace compression, $T_S$ is kept in the subspace effortlessly. In contrast, under homogeneous SparseLoCo outer synchronization, averaging across compressed and uncompressed replicas can drive the embedding buffer $T_S$ outside $\mathcal{S}$. We address this with two modifications. First, before training we initialize $T_S \leftarrow \mathrm{TE} \cdot UU^\top$ and $T_\perp \leftarrow \mathrm{TE}-T_S$, where $\mathrm{TE}$ is the standard token embedding table from uncompressed replicas. Second, only after each SparseLoCo outer synchronization, we project $T_S$ back to $\mathcal{S}$ while accumulating the projection residual into $T_\perp$: 
 
 \begin{align} 
 T_\perp &\leftarrow T_\perp + \bigl(T_S - \Pi_{\mathcal{S}}(T_S)\bigr) \\ 
 T_S &\leftarrow \Pi_{\mathcal{S}}(T_S)\, . 
 \end{align}
 
 This guarantees $T_S \in \mathcal{S}$ at the start of each inner loop, while $T_\perp$ preserves any out-of-subspace drift, ensuring $\mathrm{TE}$ is the same for all replicas after each synchronization.

\paragraph{Analysis of heterogeneous aggregation.}
Subspace projection introduces systematic bias in the pseudo-gradients. Let $\Delta^*$ denote the unbiased pseudo-gradient and $\Delta^*_{\text{proj}} = \Pi_{\mathcal{S}}(\Delta^*)$ the projected variant. The compression bias is $B = \Delta^* - \Delta^*_{\text{proj}}$.

With a fraction $\alpha$ of replicas running uncompressed, the aggregated pseudo-gradient has expected value:
\begin{equation}
\mathbb{E}[\bar{\Delta}_{\text{het}}] = \alpha \Delta^* + (1-\alpha) \Delta^*_{\text{proj}} = \Delta^* - (1-\alpha)B
\end{equation}
Under uniform compression, the bias is $\|B\|$; under heterogeneous compression, it reduces to $(1-\alpha)\|B\|$. Since $\|B\|$ grows with compression aggressiveness (larger $d/k$), this predicts the heterogeneous advantage should scale with compression ratio---consistent with Table~\ref{tab:scaling}.

Importantly, this bias reduction mechanism requires that uncompressed replicas provide unbiased gradient estimates to anchor the aggregation. Under standard AdamW without local optimization (Table~\ref{tab:adamw}), the frequent synchronization prevents compression bias from accumulating across steps, negating the heterogeneous advantage. SparseLoCo's $H$-step local optimization allows bias to compound, making the correction from uncompressed replicas more valuable.

\section{Experiments}
\label{sec:experiments}
In this section, we empirically evaluate the compression schemes proposed in Section~\ref{sec:method} through distributed training experiments on practical language modeling benchmarks. Our experimental methodology encompasses systematic ablation studies across model scales, compression ratios, and network configurations to isolate the effects of selective compression. We focus on simulations with equivalent number of steps to determine the validation loss and thereby any performance degradation incurred by Subspace compression. This is in contrast to studying a specific low-bandwidth environment, as in \cite{protocol_models}. This approach allows us to isolate the accuracy performance and communication behavior and simulate it for a variety of bandwidth settings.
We establish three principal findings: \textbf{(1)} Activation compression between pipeline stages composes with pseudo-gradient sparsification at modest cost, \textbf{(2)} Selective compression informed by interconnect bandwidth consistently outperforms compression across all workers , and \textbf{(3)} The advantage of heterogeneous configurations grows with compression aggressiveness.

\paragraph{Training Configuration.}
We train decoder-only transformers at 178M and 512M parameter LLaMA-2 model scales on large-scale pretraining corpora DCLM~\cite{dclm} and C4~\cite{c4}, employing SparseLoCo with $M=8$ replicas partitioned across 4 pipeline stages. For SparseLoCo, we use the default hyperparameters from~\cite{SparseLoCo}: a chunk size of $64 \times 64$ (i.e., 4096 elements per chunk), $\textsc{Top-}k$ with $k{=}32$ values selected per chunk (corresponding to 0.78\% density), error feedback momentum $\beta{=}0.95$, and without quantization. We use AdamW~\cite{adam,adamw} as the inner optimizer, with learning rates of $3\times 10^{-4}$ (178M) and $1\times 10^{-3}$ (512M). Each replica processes local batches of size 32, yielding an effective global batch size of 524{,}288 tokens per inner step. Unless otherwise stated, we use $H=50$ inner steps and Chinchilla-optimal~\cite{hoffmann2022training} training budgets on the DCLM dataset ~\cite{dclm}. Complete architectural specifications, training hyperparameters, and evaluation methodology are detailed in Appendix~\ref{apdx:experimental}.

We vary the subspace projection dimension $k \in \left\{ \frac{d}{8}, \frac{d}{12}, \frac{d}{24}, \frac{d}{48}, \frac{d}{96}, \frac{d}{192}, \frac{d}{384}, \frac{d}{768} \right\}$, corresponding to compression ratios, from 87.5\% to 99.87\%. For the 512M model ($d=1536$), this translates to subspace dimensions ranging from $k=192$ to $k=2$.

We compare three distinct deployment settings:

\textbf{SparseLoCo} Our primary baseline, SparseLoCo with full-precision activations are used across all pipeline stages\\
\textbf{SparseLoCo + PP-Compress} SparseLoCo with uniform activation compression applied to all worker replicas with the same stage boundaries and compression used for each replica \\
\textbf{SparseLoCo + Het-PP-Compress} SparseLoCo with Selective compression reflecting heterogeneous interconnect bandwidth. Here, a subset of replicas use pipeline compression while others are uncompressed. This selective approach enables more aggressive compression on bandwidth-constrained links while preserving full-precision communication where network capacity permits, ultimately achieving superior loss-communication tradeoffs compared to the uniform compression strategy.

\subsection{Results}

\begin{table}[t]
\centering
\caption{SparseLoCo performance with 178M- and 512M-parameter LLaMA-2 models trained under a compute-optimal token budget. \emph{SparseLoCo} denotes the baseline method without activation compression. \emph{SparseLoCo + PP-Compress} applies subspace-compressed pipeline parallelism with 4 stages to all replicas, while \emph{SparseLoCo + Het-PP-Compress (1/2)} mixes full-precision replicas with compressed ones (half of the workers each). Entries marked with $^*$ indicate that hyperparameters were tuned specifically for the compressed setting; unmarked rows reuse the baseline SparseLoCo hyperparameters.}
\label{tab:compression_across_sizes}
\begin{tabular}{l c cc|cc}
\toprule
 & \multicolumn{1}{c}{Pipeline} & \multicolumn{2}{c}{178M} & \multicolumn{2}{c}{512M} \\
Configuration 
& Compression
& Loss & Perplexity 
& Loss & Perplexity \\
\midrule
SparseLoCo
& 0\% 
& 3.07 & 21.50 
& 2.73 & 15.40 \\

SparseLoCo + PP-Compress
& 87.5\% 
& 3.14 & 23.00 
& 2.84 & 17.10 \\

SparseLoCo + PP-Compress$^*$
& 87.5\% 
& -- & -- 
& 2.82 & 16.70 \\

SparseLoCo + Het-PP-Compress (1/2)
& 87.5\% 
& 3.12 & 22.70 
& 2.82 & 16.90 \\

SparseLoCo + Het-PP-Compress (1/2)$^*$
& 87.5\% 
& -- & -- 
& 2.80 & 16.50 \\
\bottomrule
\end{tabular}
\end{table}

\begin{figure}[t]
    \centering
        \vspace{0pt}
        \centering
        \includegraphics[width=0.6\linewidth]{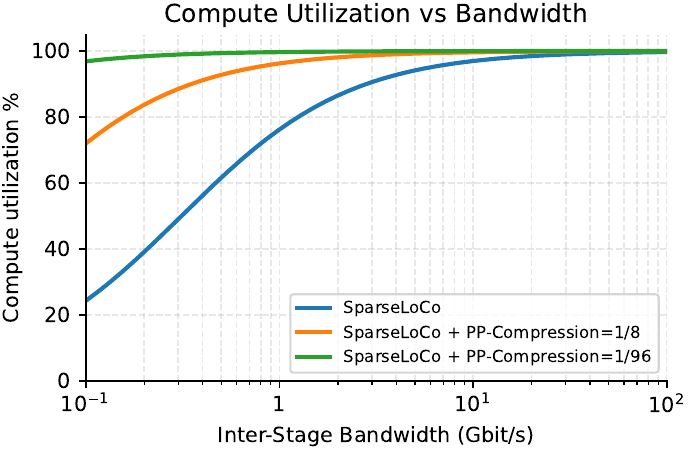}
        \caption{Compute utilization versus inter-stage bandwidth for a 70B-parameter model partitioned into 4 pipeline stages. We compare SparseLoCo with $H{=}50$ inner steps under different PP-compression ratios. 
        X-axis denotes the bandwidth between adjacent stages. 
        We observe that PP-compression significantly improves compute utilization in limited-bandwidth regimes.}
        \label{fig:compute-util}
\end{figure}

\paragraph{SparseLoCo can be practically combined with PP-Compression} Table~\ref{tab:compression_across_sizes} presents our main results on the 512M model. Here, the models are trained with the same number of inner and outer steps using the different approaches. 
We observe that (a) using our adapted version of Subspace (\textbf{PP-Compress}) compression the performance degradation is minimal compared to \textbf{SparseLoCo (Baseline)}, while allowing bandwidth limited pipelining. A modest compression of 87.5$\%$ is evaluated between pipeline stages. (b) Furthermore, we can mix replicas using pipeline compression  and those without it successfully (\textbf{Het-PP-Compress}), while actually decreasing the already small gap to the \textbf{SparseLoCo (Baseline)}. Observing Figure~\ref{fig:compute-util} which simulates this compression rate we can see that this can have significant impact on wall-clock time. Specifically, we observe that if pipeline replicas operate at even 100~\text{Mb/s} \,--\, 1~\text{Gb/s} links (a practical scenario in over the Internet settings) we can achieve higher than 97\% compute utilization while SparseLoCo naively would not support this if stage level links are low-bandwidth.

We also studied the impact of hyperparameter tuning. Note that the SparseLoCo baseline in Table~\ref{tab:compression_across_sizes} is already well tuned, we thus show results with Pipeline compression using the same hyperparameters (inner and outer learning rates). On the other hand we hypothesized that moderate hyperparameter tuning specific to PP-compression can actually improve performance, this is confirmed in the table where we are able to get better performance with tuning in the specific target bandwidth setting (which may be practical in certain scenarios). 

\begin{table}[h]
\centering
\caption{Performance of 512M models trained with SparseLoCo under different subspace compression ratios. In PP-Compress, all replicas perform PP-compression, and Het-PP-Compress, half of the workers are uncompressed. Heterogeneous setting consistently outperforms uniform PP-Compression, and the benefits grow with higher subspace compression. }
\label{tab:scaling}

{\footnotesize
\setlength{\tabcolsep}{4pt}
\begin{tabular}{@{}lcccccc@{}}
\toprule
& & \multicolumn{2}{c}{PP-Compress} & \multicolumn{2}{c}{Het-PP-Compress (1/2)} \\
\cmidrule(lr){3-4} \cmidrule(lr){5-6}
$k/d$ & Compression & Loss & $\Delta$ (\%) & Loss & $\Delta$ (\%) \\
\midrule
 1 (SparseLoCo)& 0\% & 2.73 & --- &   2.73    & --\\
\midrule
1/8   & 87.5\% & 2.84 & 3.8 & 2.82 & 3.3 \\
1/24  & 95.8\% & 2.89 & 5.8 & 2.88 & 5.4 \\
1/96  & 99.0\% & 2.96 & 8.1 & 2.94 & 7.4 \\
1/192 & 99.5\% & 3.03 & 10.9 & 2.97 & 8.8 \\
1/384 & 99.7\% & 3.04 & 11.3 & 2.99 & 9.4 \\
1/768 & 99.9\% & 3.07 & 12.4 & 3.00 & 9.8 \\
\bottomrule
\end{tabular}}
\end{table}

Our application of the pipeline compression uses a modest compression compared to \cite{protocol_models} due to the desire to maintain performance on a per-iteration basis. We study further the effect of the compression level in Table \ref{tab:scaling} which varies the pipeline compression ratio ($k/d$) for a 512M model trained with SparseLoCo. At 87.5\% compression, heterogeneous configurations gain 0.50 percentage points over uniform. At 99.9\%, where the compression error is much stronger, the heterogeneous setting is more effective with  a significantly lower loss degradation. We note that although 10\% performance degradation at these very high compression settings is significant, it can in many cases correspond to training the model for a few additional steps. Given the higher compute utilization, this can be possible within a given time. On the other hand, aggressive compression of as much as 99.9\% can unlock significant resources at lower per flop costs, accelerating the overall training and potentially reducing its cost. 

\begin{figure}[t]
    \centering
    \begin{subfigure}[t]{0.48\linewidth}
        \vspace{5pt}
        \centering
        {\scriptsize 
        \setlength{\tabcolsep}{4pt}
        \renewcommand{\arraystretch}{1.05}
        \begin{tabular}{@{}lcc@{}}
        \toprule
        Configuration & Token Budget & Loss \\
        \midrule
        SparseLoCo (Baseline) & 10B & 2.73 \\
        SparseLoCo + PP-Compress & 12B & 2.78 \\
        SparseLoCo + Het-PP-Compress (1/2) & 12B & 2.75 \\
        \bottomrule
        \end{tabular}}
        \caption{Final loss for 512M under extended token budgets. (Compressed runs use $k/d{=}1/8$).}
        \label{fig:longer_table}
    \end{subfigure}\hfill
    \begin{subfigure}[t]{0.48\linewidth}
        \vspace{0pt}
        \centering
        \includegraphics[width=\linewidth]{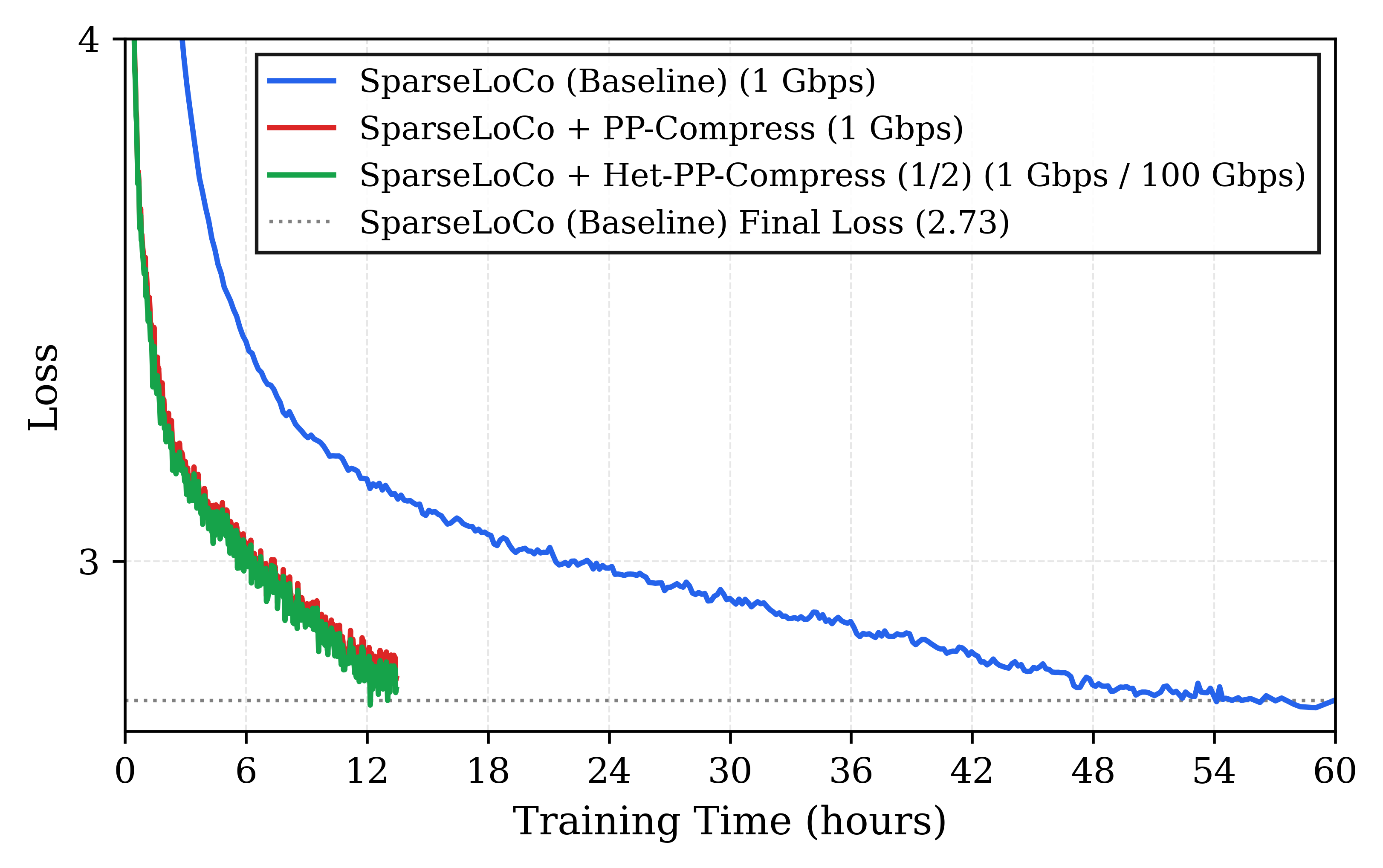}
        \caption{Simulated wall clock times for 1 Gbps inter-stage bandwidth.}
        \label{fig:wallclock_1gbps}
    \end{subfigure}

    \caption{Extended token budget results (left) and wall-clock training time under bandwidth constraints (right) for a 512M model.  }
    \label{fig:wallclock_longer}
\end{figure}

To validate this, we extend the token budget from 10B to 12B (+20\%). Table~\ref{fig:longer_table} shows that with 20\% additional training flops, heterogeneous compression achieves nearly the same performance as the baseline. Figure~\ref{fig:wallclock_1gbps} simulates the benefits at 1Gb/s cross stage links, compressed training is significantly faster than uncompressed at low-bandwidths between stages.

These results illustrate a general principle: for any bandwidth constraint, there exists a compression ratio $\frac{k}{d}$ that reduces communication overhead below the compute bottleneck. The resulting performance gap is recovered by training on additional tokens---which remains feasible within the same wall-clock budget precisely because of the reduced communication cost. With compression ratios as aggressive as $\frac{k}{d}{=}\frac{1}{768}$ yielding single-digit degradation (Table~\ref{tab:scaling}), this tradeoff is practical across a wide range of bandwidth constraints, from consumer Internet to cross-datacenter links.

\begin{table}[h]
\centering
\caption{Normal distributed data parallel (DDP) performance of 512M models trained with AdamW. PP-Compression indicates all replicas performing subspace-compressed PP and in Heterogeneous, only half of the replicas perform compression. Unlike SparseLoCo, we observe heterogeneous setting is in fact detrimental in this setting.}
\label{tab:adamw}
{\footnotesize
\begin{tabular}{lccc}
\toprule
Configuration & $k/d$ & Loss & $\Delta$ (\%) \\
\midrule
AdamW & --- & 2.75 & --- \\
AdamW + PP-Compression & 1/8 & 2.81 & 2.2 \\
AdamW + Heterogeneous PP-Compression & 1/8 & 2.83 & 2.9 \\
\bottomrule
\end{tabular}}
\end{table}

\paragraph{Heterogeneous advantage in AdamW} We also analyze whether the observation of heterogeneous replicas improving performance applies to standard AdamW training as studied in \cite{protocol_models}. Our results are shown in Table~\ref{tab:adamw}. We observe that when mimicking per step training, similar performance degradation as in SparseLoCo is observed, on the other hand the heterogeneous advantage that we have observed does not occur in the case of standard AdamW training. 

\section{Ablations and Analysis}
In this work, we modify Subspace Networks~\cite{protocol_models} by introducing token-embedding adaptation and removing components that are deemed unhelpful under SparseLoCo, namely weight projection and subspace updates using Grassmann manifolds. This section ablates these design choices and reports their impact on training loss. In all cases the same SparseLoCo baseline is used as reported in \ref{tab:compression_across_sizes} for the 512M model.
\begin{table*}[!h]
\vspace{-4pt}
\centering
\caption{Ablations on token-embedding adaptation, weight projection, Grassmann subspace adaptation, and modified AdamW, training a 512M model with SparseLoCo where all replicas perform PP-compression with $\frac{k}{d}{=}\frac{1}{8}$ compression ratio. We observe that token-embedding adaptation significantly improves the performance, while weight projection, subspace adaptation and modified AdamW are negligible.}
\label{tab:ablations_all}

\footnotesize
\setlength{\tabcolsep}{4pt}
\renewcommand{\arraystretch}{1.05}

\begin{subtable}[h]{0.48\textwidth}
\centering
\caption{Token-embedding adaptation.}
\label{tab:tfixed_ablation}
\vspace{2pt}
\begin{tabular}{lcc}
\toprule
$\mathrm{TE}$ Adaptation & Loss & $\Delta$ (\%) \\
\midrule
Without & 2.89 & 5.8 \\
With    & 2.82 & 3.3 \\
\bottomrule
\end{tabular}
\end{subtable}\hfill
\begin{subtable}[h]{0.48\textwidth}
\centering
\caption{Weight projection.}
\label{tab:wp_ablation}
\vspace{2pt}
\begin{tabular}{lccc}
\toprule
Configuration & Weight Proj. & Loss & $\Delta$ (\%) \\
\midrule
Uniform & Yes & 2.88 & 5.5 \\
Uniform & No  & 2.84 & 3.8 \\
Heterogeneous  & Yes & 2.87 & 4.9 \\
Heterogeneous & No  & 2.82 & 3.3 \\
\bottomrule
\end{tabular}
\end{subtable}

\vspace{4pt}

\begin{subtable}[h]{0.46\textwidth}
\centering
\caption{Subspace adaptation. A Grassmann update \cite{protocol_models} is applied every 500 steps.}
\label{tab:grassmann}
\vspace{2pt}
\begin{tabular}{lccc}
\toprule
Configuration & Grassmann & Loss & $\Delta$ (\%) \\
\midrule
Uniform & Yes & 2.84 & 3.8 \\
Uniform & No   & 2.84 & 3.8 \\
Heterogeneous  & Yes & 2.82 & 3.3 \\
Heterogeneous & No    & 2.82 & 3.3 \\
\bottomrule
\end{tabular}
\end{subtable}\hfill
\begin{subtable}[h]{0.46\textwidth}
\centering
\caption{Modified AdamW.}
\label{tab:modified_adamw_ablation}
\vspace{2pt}
\begin{tabular}{lccc}
\toprule
Configuration & Modif. AdamW & Loss & $\Delta$ (\%) \\
\midrule
Uniform & No  & 2.84 & 3.8 \\
Uniform & Yes & 2.84 & 3.8 \\
Heterogeneous & No  & 2.82 & 3.3 \\
Heterogeneous  & Yes & 2.82 & 3.3 \\
\bottomrule
\end{tabular}
\end{subtable}

\vspace{2pt}
\end{table*}

\paragraph{Token embedding adaptation.}In Section~\ref{sec:token}, we introduce a token-embedding adaptation for the heterogeneous setting, applied after each cross-replica update, to ensure that $T_S$ remains in the subspace $\mathcal{S}$ throughout training. Table~\ref{tab:tfixed_ablation} (a) shows that this adaptation improves optimization.

\paragraph{Weight Projection}
Subspace Networks~\cite{protocol_models} projects weight matrices $W_{p_1}$ onto $\mathcal{S}$ after each optimization step. In Table~\ref{tab:ablations_all} (b), we observe that this is counterproductive under SparseLoCo.

\paragraph{Random Subspace Suffices}

Subspace Networks~\cite{protocol_models} also proposes adapting the projection basis $U$ via optimization on the Grassmann manifold. Table~\ref{tab:ablations_all} (c) suggests that this provides no benefit in our setting.
A fixed random orthonormal basis yield similar performance, and the subspace does not benefit from dynamic adaptation.

\paragraph{Modified AdamW} A modified version of AdamW is used in~\cite{protocol_models}, it was observed in our setting with SparseLoCo that this does not yield any benefit, thus simplifying design choices.

\section{Conclusion}
In this work, we studied how to combine SparseLoCo~\cite{SparseLoCo} with a communication-efficient pipeline parallelism method to enable LLM pre-training under low-bandwidth constraints. We further considered a heterogeneous setting where well-provisioned participants run uncompressed replicas, while resource-limited participants collectively form a replica where they each act as a single stage in the pipeline. To support the heterogeneous setting, we employed subspace-compressed pipeline parallelism~\cite{protocol_models} with modifications for the heterogeneous SparseLoCo training. Across multiple model sizes and datasets, our experiments show that PP-compressed SparseLoCo introduces minimal performance degradation. The heterogeneous training further improves the PP-compressed setting, with larger gains at higher compression rates. Finally, we demonstrate that the benefit of mixing compressed, and uncompressed replicas is specific to SparseLoCo, and is in fact detrimental under standard AdamW training without inner steps.

\bibliography{ref}

@inproceedings{
protocol_models,
title={Subspace Networks: Scaling Decentralized Training with Communication-Efficient Model Parallelism},
author={Sameera Ramasinghe and Thalaiyasingam Ajanthan and Gil Avraham and Yan Zuo and Alexander Long},
booktitle={The Thirty-ninth Annual Conference on Neural Information Processing Systems},
year={2025},
url={https://openreview.net/forum?id=kke9TwtKi0}
}

@article{sparseloco,
  title={Communication Efficient LLM Pre-training with SparseLoCo},
  author={Sarfi, Amir and Th{\'e}rien, Benjamin and Lidin, Joel and Belilovsky, Eugene},
  journal={arXiv preprint arXiv:2508.15706},
  year={2025}
}

@inproceedings{local_sgd,
  title={Local SGD Converges Fast and Communicates Little},
  author={Stich, Sebastian U},
  booktitle={International Conference on Learning Representations},
  year={2019}
}

@article{demo,
  title={Decoupled momentum optimization},
  author={Peng, Bowen and Quesnelle, Jeffrey and Kingma, Diederik P},
  journal={arXiv preprint arXiv:2411.19870},
  year={2024}
}

@inproceedings{
adaptive_fed_opt,
title={Adaptive Federated Optimization},
author={Sashank J. Reddi and Zachary Charles and Manzil Zaheer and Zachary Garrett and Keith Rush and Jakub Kone{\v{c}}n{\'y} and Sanjiv Kumar and Hugh Brendan McMahan},
booktitle={International Conference on Learning Representations},
year={2021},
url={https://openreview.net/forum?id=LkFG3lB13U5}
}

@article{transformer,
  title={Attention is all you need},
  author={Vaswani, Ashish and Shazeer, Noam and Parmar, Niki and Uszkoreit, Jakob and Jones, Llion and Gomez, Aidan N and Kaiser, {\L}ukasz and Polosukhin, Illia},
  journal={Advances in neural information processing systems},
  volume={30},
  year={2017}
}

@article{llama1,
  title={Llama: Open and efficient foundation language models},
  author={Touvron, Hugo and Lavril, Thibaut and Izacard, Gautier and Martinet, Xavier and Lachaux, Marie-Anne and Lacroix, Timoth{\'e}e and Rozi{\`e}re, Baptiste and Goyal, Naman and Hambro, Eric and Azhar, Faisal and others},
  journal={arXiv preprint arXiv:2302.13971},
  year={2023}
}

@inproceedings{adam,
  title={Adam: A Method for Stochastic Optimization},
  author={Kingma, Diederik P and Ba, Jimmy},
  booktitle={International Conference on Learning Representations (ICLR)},
  year={2015},
  url={https://arxiv.org/abs/1412.6980}
}

@inproceedings{swarm,
  title={Swarm parallelism: Training large models can be surprisingly communication-efficient},
  author={Ryabinin, Max and Dettmers, Tim and Diskin, Michael and Borzunov, Alexander},
  booktitle={International Conference on Machine Learning},
  pages={29416--29440},
  year={2023},
  organization={PMLR}
}

@article{singh2025model,
  title={Model Parallelism With Subnetwork Data Parallelism},
  author={Singh, Vaibhav and Khalid, Zafir and Oyallon, Edouard and Belilovsky, Eugene},
  journal={arXiv preprint arXiv:2507.09029},
  year={2025}
}

@article{acco,
  title={ACCO: Accumulate While You Communicate for Communication-Overlapped Sharded LLM Training},
  author={Nabli, Adel and Fournier, Louis and Erbacher, Pierre and Serrano, Louis and Belilovsky, Eugene and Oyallon, Edouard},
  journal={arXiv preprint arXiv:2406.02613},
  year={2024}
}

@InProceedings{pmlr-v202-wang23t,
  title = 	 {{C}ocktail{SGD}: Fine-tuning Foundation Models over 500{M}bps Networks},
  author =       {Wang, Jue and Lu, Yucheng and Yuan, Binhang and Chen, Beidi and Liang, Percy and De Sa, Christopher and Re, Christopher and Zhang, Ce},
  booktitle = 	 {Proceedings of the 40th International Conference on Machine Learning},
  pages = 	 {36058--36076},
  year = 	 {2023},
  editor = 	 {Krause, Andreas and Brunskill, Emma and Cho, Kyunghyun and Engelhardt, Barbara and Sabato, Sivan and Scarlett, Jonathan},
  volume = 	 {202},
  series = 	 {Proceedings of Machine Learning Research},
  month = 	 {23--29 Jul},
  publisher =    {PMLR}
}

@inproceedings{adamw,
  author       = {Ilya Loshchilov and
                  Frank Hutter},
  title        = {Decoupled Weight Decay Regularization},
  booktitle    = {7th International Conference on Learning Representations, {ICLR} 2019,
                  New Orleans, LA, USA, May 6-9, 2019},
  publisher    = {OpenReview.net},
  year         = {2019}}

@inproceedings{
douillard2025streaming,
title={Streaming DiLoCo with overlapping communication},
author={Arthur Douillard and Yani Donchev and J Keith Rush and Satyen Kale and Zachary Charles and Gabriel Teston and Zachary Garrett and Jiajun Shen and Ross McIlroy and David Lacey and Alexandre Rame and Arthur Szlam and MarcAurelio Ranzato and Paul R Barham},
booktitle={Second Conference on Language Modeling},
year={2025},
url={https://openreview.net/forum?id=yYk3zK0X6Q}
}

@misc{muloco,
  author       = {Benjamin Th{\'{e}}rien and
                  Xiaolong Huang and
                  Irina Rish and
                  Eugene Belilovsky},
  title        = {MuLoCo: Muon is a practical inner optimizer for DiLoCo},
  journal      = {CoRR},
  volume       = {abs/2505.23725},
  year         = {2025},
  url          = {https://arxiv.org/abs/2505.23725},
}

@article{diloco,
  author       = {Arthur Douillard and
                  Qixuang Feng and
                  Andrei A. Rusu and
                  Rachita Chhaparia and
                  Yani Donchev and
                  Adhiguna Kuncoro and
                  Marc'Aurelio Ranzato and
                  Arthur Szlam and
                  Jiajun Shen},
  title        = {DiLoCo: Distributed Low-Communication Training of Language Models},
  journal      = {CoRR},
  volume       = {abs/2311.08105},
  year         = {2023},
  url          = {https://doi.org/10.48550/arXiv.2311.08105},}

@article{hoffmann2022training,
  title={Training compute-optimal large language models},
  author={Hoffmann, Jordan and Borgeaud, Sebastian and Mensch, Arthur and Buchatskaya, Elena and Cai, Trevor and Rutherford, Eliza and Casas, Diego de Las and Hendricks, Lisa Anne and Welbl, Johannes and Clark, Aidan and others},
  journal={arXiv preprint arXiv:2203.15556},
  year={2022}
}

@inproceedings{dclm,
  author       = {Jeffrey Li and
                  Alex Fang and
                  Georgios Smyrnis and
                  Maor Ivgi and
                  Matt Jordan and
                  Samir Yitzhak Gadre and
                  Hritik Bansal and
                  Etash Kumar Guha and
                  Sedrick Scott Keh and
                  Kushal Arora and
                  Saurabh Garg and
                  Rui Xin and
                  Niklas Muennighoff and
                  Reinhard Heckel and
                  Jean Mercat and
                  Mayee F. Chen and
                  Suchin Gururangan and
                  Mitchell Wortsman and
                  Alon Albalak and
                  Yonatan Bitton and
                  Marianna Nezhurina and
                  Amro Abbas and
                  Cheng{-}Yu Hsieh and
                  Dhruba Ghosh and
                  Josh Gardner and
                  Maciej Kilian and
                  Hanlin Zhang and
                  Rulin Shao and
                  Sarah M. Pratt and
                  Sunny Sanyal and
                  Gabriel Ilharco and
                  Giannis Daras and
                  Kalyani Marathe and
                  Aaron Gokaslan and
                  Jieyu Zhang and
                  Khyathi Raghavi Chandu and
                  Thao Nguyen and
                  Igor Vasiljevic and
                  Sham M. Kakade and
                  Shuran Song and
                  Sujay Sanghavi and
                  Fartash Faghri and
                  Sewoong Oh and
                  Luke Zettlemoyer and
                  Kyle Lo and
                  Alaaeldin El{-}Nouby and
                  Hadi Pouransari and
                  Alexander Toshev and
                  Stephanie Wang and
                  Dirk Groeneveld and
                  Luca Soldaini and
                  Pang Wei Koh and
                  Jenia Jitsev and
                  Thomas Kollar and
                  Alex Dimakis and
                  Yair Carmon and
                  Achal Dave and
                  Ludwig Schmidt and
                  Vaishaal Shankar},
  editor       = {Amir Globersons and
                  Lester Mackey and
                  Danielle Belgrave and
                  Angela Fan and
                  Ulrich Paquet and
                  Jakub M. Tomczak and
                  Cheng Zhang},
  title        = {DataComp-LM: In search of the next generation of training sets for
                  language models},
  booktitle    = {Advances in Neural Information Processing Systems 38: Annual Conference
                  on Neural Information Processing Systems 2024, NeurIPS 2024, Vancouver,
                  BC, Canada, December 10 - 15, 2024},
  year         = {2024},
  url          = {http://papers.nips.cc/paper\_files/paper/2024/hash/19e4ea30dded58259665db375885e412-Abstract-Datasets\_and\_Benchmarks\_Track.html},
 }

@article{swiglu,
  title={Glu variants improve transformer},
  author={Shazeer, Noam},
  journal={arXiv preprint arXiv:2002.05202},
  year={2020}
}

@article{c4,
  title={Exploring the limits of transfer learning with a unified text-to-text transformer},
  author={Raffel, Colin and Shazeer, Noam and Roberts, Adam and Lee, Katherine and Narang, Sharan and Matena, Michael and Zhou, Yanqi and Li, Wei and Liu, Peter J},
  journal={Journal of machine learning research},
  volume={21},
  number={140},
  pages={1--67},
  year={2020}
}
\bibliographystyle{abbrv}

\clearpage
\appendix
\section{Experiment setup}
\label{apdx:experimental}

\paragraph{Model Architecture.} We train decoder-only transformers following the LLaMA architecture~\cite{llama1} with SwiGLU activations~\cite{swiglu} at three scales: 178M, 512M and 1B parameters. Architectural specifications are provided in Table~\ref{tab:models}.

\begin{table}[h] \centering \caption{Model architectural configurations and training corpus sizes.} \label{tab:models} \begin{tabular}{lccccc} \toprule Parameters & Hidden Dim ($d$) & Layers & Attention Heads & Context Length & Training Tokens \\
\midrule 
178M & 1024 & 9 & 8 & 2048 & 3B \\ 
512M & 1536 & 12 & 12 & 2048 & 10B \\ 
1B & 1536 & 16 & 8 & 1024 & 10B \\ 
\bottomrule 
\end{tabular} 
\end{table}

Training loss is the average loss across data parallel replicas, evaluated on unseen data at each outer optimization step.
Relative performance degradation is defined as $\Delta=\frac{\mathcal{L}_{\text{compressed}} - \mathcal{L}_{\text{baseline}}}{\mathcal{L}_{\text{baseline}}} \times 100\%$

\paragraph{Scaling to 1B parameters.}
Table~\ref{tab:1b} confirms that PP-compression remains effective at the 1B scale, with 87.5\% compression yielding a loss of 2.910 compared to the baseline 2.747---a degradation consistent with smaller model scales.

\begin{table}[h]
\centering
\caption{1B parameter results with SparseLoCo on DCLM (sequence length 1024, 10B tokens).}
\label{tab:1b}
\begin{tabular}{lcc}
\toprule
Configuration & Compression & Loss \\
\midrule
Baseline & 0\% & 2.747 \\
PP-Compression & 87.5\% & 2.910 \\
\bottomrule
\end{tabular}
\end{table}

\paragraph{Generalization to C4 dataset.}
Table~\ref{tab:c4} reports results for a 512M model trained for 10B tokens on the C4 dataset ~\cite{c4}. The trends mirror those on DCLM: PP-compression introduces modest degradation (3.71\%), and the heterogeneous configuration improves over uniform compression (3.32\% vs 3.71\%). This demonstrates that the benefits of heterogeneous training generalize across pretraining corpora.

\begin{table}[h]
\centering
\caption{512M parameter results with SparseLoCo on C4 (sequence length 2048, 10B tokens)}
\label{tab:c4}
{\footnotesize
\begin{tabular}{lccc}
\toprule
Configuration & $k/d$ & Loss & $\Delta$ (\%) \\
\midrule
Baseline & --- & 2.67 & --- \\
PP-Compression & 1/8 & 2.77 & 3.71 \\
Heterogeneous PP-Compression & 1/8 & 2.76 & 3.32 \\
\bottomrule
\end{tabular}}
\end{table}

\begin{table}[htbp]
\centering
\caption{Hyperparameters and model architecture used for training.}
\label{tab:hyperparameters}
\begin{tabular}{lr}
\toprule
\textbf{Parameter}                  & \textbf{Value}                  \\
\midrule
Model size (parameters)             & 178M / 512M / 1B                  \\
Architecture                        & Decoder-only LLaMa Transformer        \\
Number of layers                    & 9 / 12 / 16                    \\
Hidden size ($d_{model}$)           & 1024 / 1536 / 2048               \\
Number of attention heads           & 8 / 12 / 8                    \\
Head dimension                      & 128                             \\
FFN intermediate size               & $\{2.63 / 2.54 / 4\} \times d_{model}$             \\
FFN Activation                      & SwiGLU \\
Vocabulary size                     & 32K        \\
Tokenizer                           & Byte-Pair Encoding (BPE) / SentencePiece \\
Context length                      & 2048 / 2048 / 1024 tokens       \\
Batch size (tokens)                 & 512K / 512K / 65K                  \\
Training tokens                     & 3B / 10B / 10B                         \\
Training steps                      & 5.8K / 19.5K / 156.5K                       \\
Learning rate schedule              & Cosine decay                    \\
Warmup steps                        & 500         \\
Number of replicas ($M$)            & 8 / 8 / 2                       \\
\midrule
\textbf{AdamW Optimizer}            &  \\
\midrule
$\beta_1$, $\beta_2$                & 0.9, 0.95                   \\
Weight decay                        & 0.1                       \\
\midrule
\textbf{AdamW Baseline ($H=1$)}     &  \\
\midrule
Learning rate                       & 3e-4              \\
\midrule
\textbf{SparseLoCo ($H=50$)}        &  \\
\midrule
\textit{Inner optimizer}            & AdamW (see above) \\
Inner learning rate                 & 1e-3              \\
Gradient clipping                   & 1.0                             \\
\textit{Outer optimizer}            & SGD (no momentum) \\
Outer learning rate                 & 1.0              \\
\textit{Compression}                &  \\
\quad Chunk size                    & $64 \times 64$ (4096) \\
\quad $\textsc{Top-}k$ per chunk    & 32 \\
\quad Error feedback momentum ($\beta$) & 0.95 \\
Outer steps                         & 116 / 391 / 3129                       \\
\midrule
\textbf{SparseLoCo Tuned (512M model)}    &  \\
\midrule
Warmup steps                        & 3910        \\
Outer learning rate                 & 0.8              \\
\bottomrule
\end{tabular}
\end{table}

\end{document}